\pdfoutput=1

\documentclass[11pt]{article}

\usepackage[table]{xcolor}
\usepackage[preprint]{acl}

\usepackage{times}
\usepackage{latexsym}

\usepackage[T1]{fontenc}

\usepackage[utf8]{inputenc}

\usepackage{microtype}

\usepackage{inconsolata}

\usepackage{graphicx}

\usepackage{amsmath} 
\usepackage{color}
\usepackage{colortbl}

\usepackage{xcolor}
\newcommand{\red}[1]{\textcolor{red}{#1}}
\usepackage{multirow}

\title{Enhancing Confidence Expression in Large Language Models Through Learning from Past Experience}

\author {
    Haixia Han\textsuperscript{\rm 1},
    Tingyun Li \textsuperscript{\rm 2}\thanks{Equal Contribution},
    Shisong Chen\textsuperscript{\rm 1},
    Jie Shi\textsuperscript{\rm 3},
    Chengyu Du \textsuperscript{\rm 3},\\
    \textbf{Yanghua Xiao} \textsuperscript{\rm 1,3 }\thanks{Corresponding Authors},
    \textbf{Jiaqing Liang} \textsuperscript{\rm 2$\dagger$},
    \textbf{Xin Lin} \textsuperscript{\rm 1}
    \\
    Shanghai Institute of Artificial Intelligence for Education, East China Normal University\textsuperscript{\rm 1}\\
    Shanghai Key Laboratory of Data Science, School of Data Science, Fudan University\textsuperscript{\rm 2}\\
    Shanghai Key Laboratory of Data Science, School of Computer Science, Fudan University\textsuperscript{\rm 3}\\
    \texttt{\{haixiahan03,litinyun0715,cheniison2015,cydu2024\}@gmail.com} \\
    \texttt{jshi22@m.fudan.edu.cn}, \texttt{\{liangjiaqing,shawyh\}@fudan.edu.cn}, xlin@cs.ecnu.edu.cn
}

\begin{document}
\maketitle
\begin{abstract}          
Large Language Models (LLMs) have exhibited remarkable performance across various downstream tasks, but they may generate inaccurate or false information with a confident tone. One of the possible solutions is to empower the LLM confidence expression capability, in which the confidence expressed can be well-aligned with the true probability of the generated answer being correct. However, leveraging the intrinsic ability of LLMs or the signals from the output logits of answers proves challenging in accurately capturing the response uncertainty in LLMs. Therefore, drawing inspiration from cognitive diagnostics, we propose a method of \underline{Le}arning from \underline{P}ast \underline{e}xperience (LePe) to enhance the capability for confidence expression. Specifically, we first identify three key problems: \textbf{ (1) How to capture the inherent confidence of the LLM? (2) How to teach the LLM to express confidence? (3) How to evaluate the confidence expression of the LLM?} Then we devise three stages in LePe to deal with these problems. Besides, to accurately capture the confidence of an LLM when constructing the training data, we design a complete pipeline including question preparation and answer sampling. We also conduct experiments using the Llama family of LLMs to verify the effectiveness of our proposed method on four datasets.
\end{abstract}
\section{Introduction}

\begin{figure}
\centering
\includegraphics[width=0.5\textwidth]{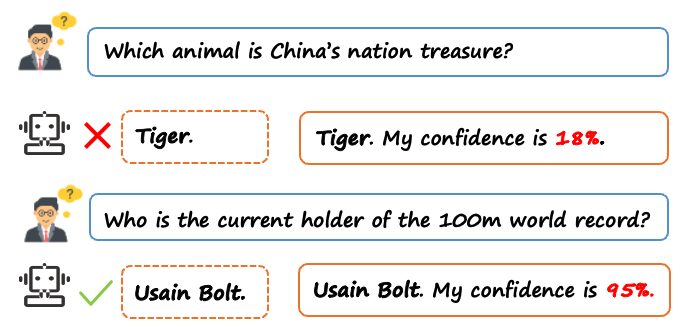}
\caption{For a given question, the LLM simply provides an answer and we can't identify the correctness. After empowering the confidence expression ability of the LLM, we can determine the trustworthiness of its responses based on the confidence level it provides. }
\label{fig:fig_1}
\end{figure}

Although large language models (LLMs) demonstrate remarkable performance across many domains \cite{guo2023close, Han2024SmallLM, achiam2023gpt}, they can't provide a reasonable confidence level in the generated answers \cite{wang2022self,shuster2021retrieval}, which distinguish them from human intelligence. A critical aspect of human intelligence is the capability to express confidence effectively.   Reliable uncertainty estimates are also vital for human-machine collaboration, offering valuable insights into response reliability and alleviating hallucinations in natural language generation (NLP) tasks \cite{xiong2023can}. Moreover, these estimates aid developers in pinpointing LLM weaknesses for targeted refinement, thus enhancing overall performance iteratively. Figure \ref{fig:fig_1} illustrates the difference in responses given by an LLM when it answers a question incorrectly or accurately, compared to the output from an LLM that can convey confidence level in its responses.

However, leveraging the intrinsic ability of LLMs or the signals from the output logits of answers proves challenging in accurately capturing the uncertainty associated with the responses generated by LLMs. Some works utilize carefully constructed prompts to instruct and generate answers while providing a confidence level \cite{lin2022teaching}. However, when verbalizing their confidence, LLMs tend to exhibit high confidence \cite{xiong2023can}. Furthermore, their effectiveness is often constrained by the task-specific nature and the labor-intensive process of designing prompts. Another method is to use the logit value of a specific token within the generated answer as a measure of uncertainty for the entire response \cite{Kadavath2022LanguageM}, such as the final numerical result in a mathematical question or the final option in a multiple-choice question \cite{li2024think}. Nevertheless, relying solely on a specific token logit makes it challenging to represent the uncertainty level of the entire generated answer accurately.

We argue that it is essential to explicitly train the LLM to express confidence, which is regarded as a meta-capability in this paper. Therefore, guided by the \emph{Cognitive Diagnostic} \cite{liu2024survey} approach for assessing students' ability levels, we propose a method of \underline{Le}arning from \underline{P}ast \underline{e}xperience (LePe) to enhance the LLM’s capability of confidence expression. In Cognitive Diagnosis, student knowledge mastery is modeled by analyzing their performance on prior experiences, thereby enabling an accurate assessment of their degree of knowledge acquisition and their potential performance on similar problems in the future. Similarly, our proposed method LePe mainly includes three stages: testing, learning, and predicting. The testing stage primarily aims to capture the inherent confidence of LLM by assessing its performance across a predefined set of questions. In the learning stage, the LLM is fine-tuned using a curated set of instruction data to learn to express its confidence level. Finally, the prediction stage involves the LLM applying its newly acquired ability to express confidence when addressing new, unseen questions.

However, it is challenging to obtain accurate confidence scores of LLMs due to the context sensitivity \cite{giallanza2024contextsensitive}, where the LLM generation results are inconsistent when the same question is presented in different contexts. Therefore, we design a complete pipeline and several strategies to alleviate this issue from multiple aspects, including \emph{mutation questions} and \emph{hybrid sampling} strategies. The mutation questions is to make various transformations of the questions and options without changing the original questions to test the robustness of the answers generated by LLM. The hybrid sampling strategy uses multiple sampling methods to obtain the model's intrinsic beliefs. 

In summary, our contributions are as follows:
\begin{itemize}
\item Inspired by the approach used in cognitive diagnostics to test student’s knowledge level, we innovatively propose a method called learning from past experiences (LePe) to empower LLMs with confidence expression capability.
\item We devise a complete pipeline to capture the inherent true responses of the LLM and alleviate the generation bias.
\item We conduct experiments using the open-source family of LLMs on several datasets to validate the effectiveness of our proposed method. The experimental results show that our method enables LLMs to give reliable confidence scores that reflect the correctness of their responses.
\end{itemize}

\begin{figure*}
\centering
\includegraphics[width=0.8\textwidth]{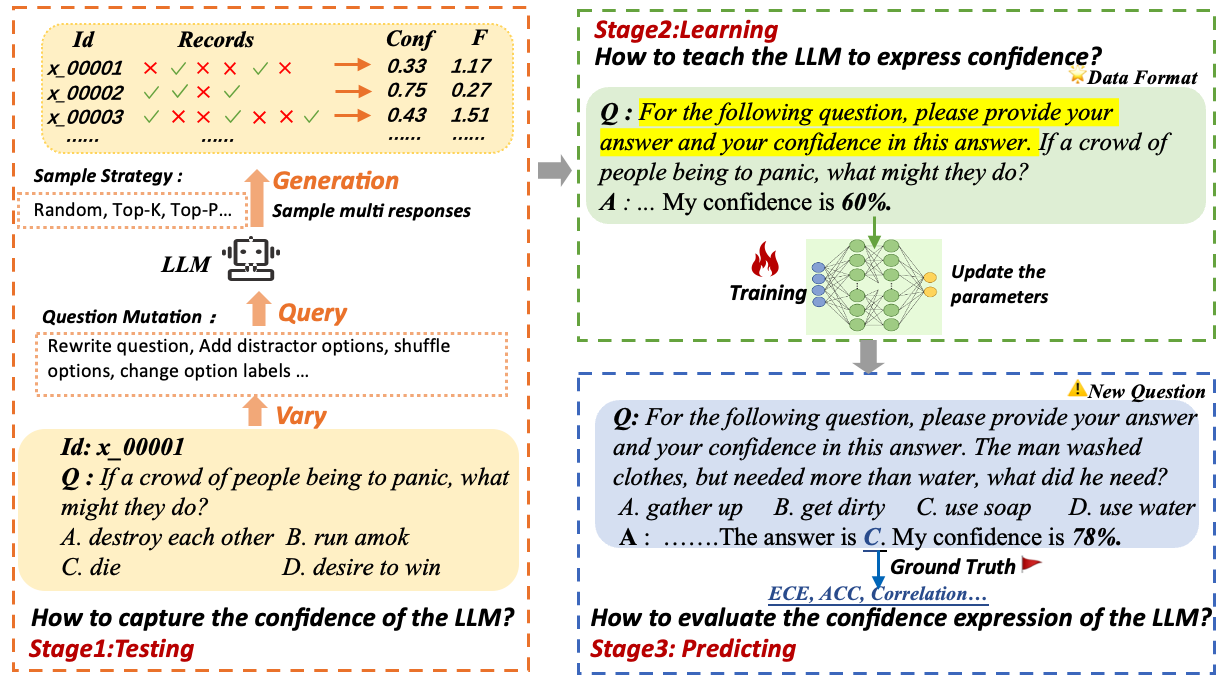}
\caption{The pipeline of our proposed method LePe.}
\label{fig:pipline}
\end{figure*}

\section{Related Work}
\subsection{Self-awareness of LLMs}  
While models were equipped with extensive parametric knowledge, some studies highlighted an evident lack of self-awareness in discerning their competence scope \cite{Wang2024MyAI}. In the existing literature, research exploring self-awareness in LLMs tended to focus on mapping the knowledge boundaries of the LLMs \cite{yin2023large,ren2023investigating}. These approaches worked to enable LLMs to make full use of their intrinsic knowledge, thereby reducing their hallucination about unknown questions. The Inference-Time Intervention (ITI) \cite{li2024inference} method worked by shifting model activations alongside factuality-related heads during inference, thereby enabling the model to generate more truthful answers. Meanwhile, the FactTune method \cite{tian2023fine} employed Direct Preference Optimization (DPO) \cite{rafailov2024direct} as a means to steer the LLM toward generating responses that aligned with external knowledge. Similarly, Srivastava et al. \citeyearpar{srivastava2022beyond} evaluated LLMs' competence in delineating their knowledge boundaries by employing a set of 23 pairs of answerable and unanswerable multiple-choice questions. However, these existing methods tended to be overly strict and conservative. When faced with uncertain questions, the LLM opted not to reply at all rather than attempting to deduce from existing information or speculate on a potential answer, thus diminishing its utility. Confidence elicitation can mitigate this issue, enabling LLMs to generate answers while conveying their confidence levels. 

\subsection{Confidence elicitation in LLMs} 
Confidence elicitation refers to the process of estimating the level of confidence in an LLM response without relying on LLM fine-tuning or access to the proprietary information of LLMs \cite{xiong2023can}. To successfully elicit the confidence of LLMs, efforts could be categorized into two categories: One focused on employing meticulously crafted prompts to guide answer generation while simultaneously eliciting confidence. Branwen \citeyearpar{Branwen} displayed GPT-3's capacity to convey expression uncertainty on basic simple queries through the few-shot prompt. This denoted the beginning of this type of approach. Lin et al. \citeyearpar{lin2022teaching} introduced the concept of "verbalized confidence", which could directly guide the LLM output confidence. Subsequently, Xiong et al. \citeyearpar{xiong2023can} aimed to explore a broader method space, introducing two categories: consistency-based methods and their hybrid variants. Zhou et al. \citeyearpar{zhou2023navigating} injected uncertainty expressions into the prompt in the hope that the LLM would emit its uncertainty expressions, but this led to a decrease in accuracy. To assess the confidence and uncertainty of language models, Tian et al. \citeyearpar{tian2023just} employed external annotations by instructing the LLM to express uncertainty in words during answer generation. However, this class of methods did not cope well with reasoning-heavy problems, and LLMs tended to be overconfident in their expression. The other category focused on using the logit value of a specific token in the generated answer to measure the uncertainty of the whole answer. Kadavath et al. \citeyearpar{Kadavath2022LanguageM} proposed probing the self-awareness of LLMs by incorporating a dedicated "Value Head". However, the effectiveness of this method faces challenges when employed in other tasks as it relies on task-specific training. 

Overall, current methods utilize the inherent capability or signal of LLMs to guide them to elicit confidence. These methods rely more on the capabilities of the model itself, and the confidence expression ability is limited by the downstream tasks. In contrast, in this paper, we regard the ability to express confidence as a meta-capability that requires explicit training in LLMs.

\section{Methods}
In this section, we first introduce our proposed LePe to empower the LLM with the confidence expression capability. Next, we provide in detail the complete pipeline for obtaining confidence scores from an LLM to construct training data. The overall framework of LePe is shown in Fig \ref{fig:pipline}. 

\begin{table*}[t]
\centering
\resizebox{\linewidth}{!}{ 
\begin{tabular}{l}
\hline
\textbf{The original question:} Sammy wanted to go to where the people were.  Where might he go?\\
A. race track \hspace{1cm} 
B. populated areas \hspace{1cm} 
C. the desert \hspace{1cm} 
D. apartment \\
\textbf{The varied question:}
\red{\emph{(TaskP)}} Examine the following options carefully and select the correct one.
\red{\emph{(COTP)}} Please select the correct \\option from the provided choices and offer a comprehensive problem-solving process. 
\red{\emph{(question)}}Sammy wanted to go to\\ where the people were.  Where might he go?\\
\red{\emph{(Shuffle options and change option label)}} 
1. populated areas \hspace{1cm} 
2. apartment \hspace{1cm} 
3. the desert \hspace{1cm} 
4. race track \\
\red{\emph{(DisO)}} 
5. All of the above / None of the above \\
\hline
\textbf{Input:} \red{\emph{(confidence expression prompt)}} For the following question, please select the correct option, and provide your confidence in this answer.\\
Google Maps and other highway and street GPS services have replaced what?\\
A. united states  \hspace{1cm} 
B. mexico  \hspace{1cm} 
C. countryside  \hspace{1cm} 
D. atlas \hspace{1cm}  E.None of the above \\
\textbf{Output:} The correct answer is \textbf{D. atlas}.  
\red{\emph{(Conf)}}\colorbox{-blue}{My confidence is 61.5\%.}\\
\hline
\end{tabular}
}
\caption{The question format of the testing stage and instruction data format of the prediction stage. }
\label{tab:format}
\end{table*}
\subsection{Learning from the Past Experience (LePe) }

In this paper, our goal is mainly to enhance the LLMs’ capability for confidence expression. 
This entails equipping the LLMs to not only generate responses but also to provide the associated confidence levels of their outputs. 
We begin by pinpointing three essential questions to enhance the confidence elicitation capability of LLMs. \textbf{(1) How to capture the inherent confidence of the LLM?} Since different LLMs demonstrate varied proficiency levels within the same knowledge domain, it's necessary to devise a standardized procedure to capture the inherent confidence of LLMs. \textbf{(2) How to teach the LLM to elicit confidence?} After gathering confidence scores of the LLM, it becomes crucial to investigate effective strategies that enable the models to convey their confidence levels. \textbf{(3) How to evaluate the confidence elicitation of the LLM?} A comprehensive assessment of LLMs' confidence elicitation abilities is required.

Inspired by the cognitive diagnostics approach, we propose the Learning confidence expression from Past Experience (LePe). Our method mainly includes three stages: testing, learning, and predicting stage. 

In the testing stage, we capture the inherent confidence level of the LLM by evaluating its previous performance. Current LLMs often exhibit high confidence when asked directly about the confidence level of their responses. To more accurately capture the inherent confidence, we evaluate the model's knowledge mastery by posing questions multiple times in different contexts. Given a question set $Q = \left \{q_{1}, \dots, q_{i}, \dots, q_{n}\right \}$, where $n$ represents the number of questions in the question set $Q$. For each question $q_i$, the LLM $M$ provides multiple corresponding answers $\left \{a_{i1}, a_{i2}, \dots, a_{ik}\right \}$, where $k$ is the total number of times the $i$th question is answered. Each answer $a_{ij}(1\le j\le k)$ is represented by $M(q_{i})\to a_{ij}$. Therefore, each question and its corresponding answer is represented as a triple tuple $(q_{i}, a_{ij}, p_{i})$, where $p_{i} (p_{i}\in\{0,1\})$ indicates the correctness of an answer $a_{ij}$, where 1 stands for correct and 0 for incorrect. Then we construct the training dataset based on these answer records. A more comprehensive discussion of this stage will be presented in the next section. 

In the learning stage, we construct the instructional data derived from the gathered confidence scores and employ instruction fine-tuning \cite{stiennon2020learning,ouyang2022training} to enhance the model's capability to convey its confidence reasonably.

In the predicting stage, we verify the calibration ability of LLM after instruction fine-tuning on the new question data by comparing the confidence of the LLM’s predictions with the probability of being correct for the answers it generates, the LLM is considered well-calibrated when its confidence estimate closely aligns with the actual probability of the answer being correct. An ideally calibrated LLM conforms to the conditions that:
\begin{equation}
\label{eq:eq4}
P(\hat{y}=y\mid conf=z)=z,\forall z\in [0,1],
\end{equation}
where $y$ is the ground truth, $\hat{y}$ is the response of the $M$, and $conf$ is the confidence level of the LLM output on this problem.

\subsection{Training Data Construction}
Traditional deep learning approaches for classification fail to capture the model uncertainty. The predictive probabilities provided by the softmax output are frequently misinterpreted as a measure of the model's confidence. However, a model may still be uncertain in its predictions despite producing a high softmax output \cite{gal2016dropout}. Therefore, we test an LLM multiple times on a batch of questions to obtain its true responses and record its performance. In this part, we also devise some strategies including \textbf{mutation question} and \textbf{answer sampling} to ensure the consistency of LLM‘s responses, alleviating context sensitivity, and capturing a more true response of LLM.\\  
\begin{figure*}
\centering
\includegraphics[width=0.8\textwidth]{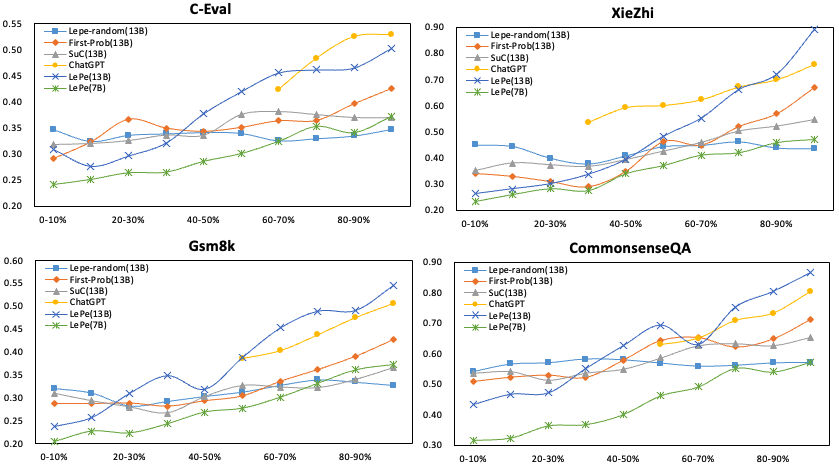}
\caption{The calibration results of the baselines. The horizontal axis is the predicted confidence and the vertical axis is the true correctness.}
\label{fig:Result}
\end{figure*}

\subsubsection{Mutation Question}
When selecting a question for the LLM to answer, we design a \emph{two-step question sampling strategy}. Firstly, each time, we randomly draw a question using sampling with the replacement for the $M$ to answer, with the total number of samplings being $k*n$. Secondly, we locate and select questions from the record that exhibit higher variability for additional inquiry. For example, in the case of a multiple-choice dataset, we assign scores to different options. For a five-option multiple-choice question, the range of scores allocated is from 1 to 5. We calculate the fuzziness $F$ of the LLM’s answer to a question: 
\begin{equation}\label{eq:eq1}
F=\frac{\sum_{i}^{k}(x_{i}-\bar{x})}{k},
\end{equation}
where $x_{i}$ is the score corresponding to the $ith$ answer of the LLM, and $\bar{x}$ is the average score of the $k$ answers. And for the math question, the fuzziness is calculated by:
\begin{equation}\label{eq:eq2}
F=\frac{u}{k},
\end{equation}
where $u$ is the number of different answers among $k$ responses. A greater fuzziness value suggests that the LLM produces various responses to the query. To obtain the model's genuine response to highly ambiguous questions, we continue querying the LLM to obtain more answer records.

To reduce the likelihood of probabilistic errors in generative models and achieve more consistent responses of LLMs, we perform various mutations to both the question stems and options before having the model answer an original question in the question set. For the question stem, we use GPT3.5-turbo to assist in \emph{rewriting the question} stem without changing the semantics. For the options, we first add some \emph{distractor options (DisO)} for multiple choice questions, such as "None of the above" or "All of the above". Besides, the options for each question are \emph{randomly shuffled (RS)} before presenting to the LLM each time. Additionally, we employ various types of option labels, including uppercase letters(A B C D...), lowercase letters (a b c d...), Arabic numerals (1 2 3 4...), as well as Roman numerals (i ii iii iv or I II III IV...). What's more, we design multiple instruction templates for a given task to guide the LLM $M$ to generate a problem-solving process, including a few-shot prompt template and \emph{a COT prompt (COTP)} template \cite{Wei2022ChainOT}. The COT prompt is like “For the given math problem, please select the correct option from the provided choices and give a comprehensive problem-solving process.” Here, we give a question example when it is presented to the LLM, which is shown in Table \ref{tab:format}. 

\subsubsection{Answer Sampling}
We use random sampling decoding when LLM infers these question sets. Besides, we have integrated the Top-k and Top-p sampling as a static solution to capture the region of confidence to avoid the unreliable tail. Combined with the prepared questions,  we obtain the LLM’s answer records on the question set $Q$. 

\begin{figure*}
\centering
\includegraphics[width=0.9\textwidth]{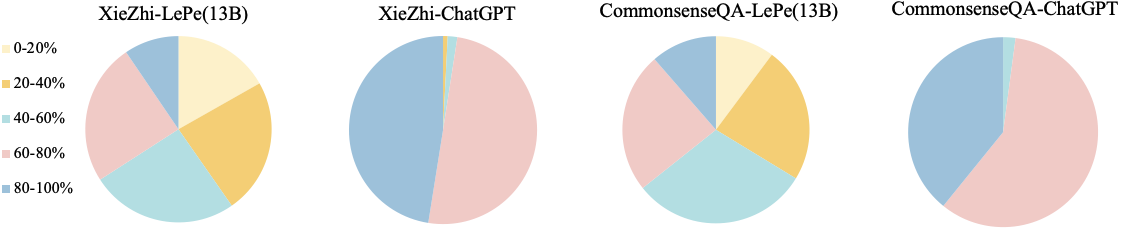}
 \caption{The detailed confidence statistics of the CuteGPT-13B and ChatGPT.}
\label{fig:pineRes}
\end{figure*}

\begin{table*}
\centering
\resizebox{0.8\linewidth}{!}{
\begin{tabular}{c >{\columncolor{blue!5}}c>{\columncolor{blue!5}}c >{\columncolor{green!10}}c >{\columncolor{green!10}}c c>{\columncolor{orange!15}}c >{\columncolor{orange!15}} c >{\columncolor{pink!20}} c >{\columncolor{pink!20}}c} 

     \hline
     \multirow{3}{*}{\textbf{Baselines}}&\multicolumn{4}{c}{\textbf{CuteGPT-13B}} & &\multicolumn{4}{c}{\textbf{LLaMA2-Chat-13B}}\\
     
     \cline{2-10} 
     
     &\multicolumn{2}{c}{C-Eval} & \multicolumn{2}{c}{XieZhi} & &  \multicolumn{2}{c}{GSM8K} & \multicolumn{2}{c}{CommonsenseQA} \\

     \cline{2-5} \cline{7-10}
     \rowcolor{white}
     &$r$ &$ECE$ &$r$ &$ECE$    
     & &$r$ &$ECE$ &$r$ &$ECE$ \\
     
     \hline
     First-Prob   &0.86 &23.36 &0.87 &15.95 & &0.92 &23.35 &0.82 &22.39\\
    \hline
     SuC          &0.90 &24.21 &0.86 &19.85 & &0.86 &26.82 &0.90 &21.79 \\
     \hline
     LePe-random  &-0.17&27.86 &0.33 &22.63 & &0.05 &27.09 &-0.20 &25.39\\
     \hline
     ChatGPT      &1.00 &21.57 &0.99 &13.08 & &0.99 &26.76 &0.98 &7.46\\
     \hline
     LePe         &0.89 &17.43 &0.98 &7.84  & &0.95 &18.77 &0.98 &14.74\\
     \hline
\end{tabular}}
\caption{The quantitative results on four datasets.}
\label{Result}
\end{table*}
\subsubsection{Data Format}

To improve the confidence expression capability of LLM, we employ the instruction fine-tuning technology commonly used in LLM capability enhancement. In the record of responses, for the $k$ responses to each question, there are three possible situations: all answers are correct, some are partially correct, or all are incorrect. Training the model with incorrect answers and low confidence levels enables the LLM to express the confidence level appropriately. However, the incorrect answers has the potential to cause the LLM to generate wrong responses, intensifying the issue of hallucinations \cite{Huang2023ASO} within the LLM. For the first two cases, we employ the following formula:
\begin{equation}\label{eq:eq3}
Conf=\frac{f_{qi}}{k},
\end{equation}
where $f_{q_{i}}$ is the number of correct answers to question $q_{i}$. We append the confidence score output after the correct answer, using a structure such as "My confidence is $\left[Conf \ast 100\right]$\%". For the last case, we incorporate incorrect answers into the training dataset. However, this content is excluded from the loss calculation, and the confidence level is set to 0. Therefore, we construct the data format like \emph{<Question, Answer+ Confidence>}. Besides, we design the confidence expression prompt. For instance, "For the following question, please provide your answer and your confidence about this answer." We further use ChatGPT to help rephrase this prompt to increase diversity and ensure LLM to better understand the instruction. The instruction data example is shown in Table \ref{tab:format}. 

\section{Experiments}
\subsection{Experiment settings}
\textbf{Dataset}. We evaluate the effectiveness of LePe on four datasets including C-Eval \cite{huang2024ceval}, XieZhi \cite{gu2024xiezhi}, GSM8K \cite{cobbe2021gsm8k}, and CommonsenseQA \cite{talmor2018commonsenseqa}. \\
\begin{table}[t]
\renewcommand{\arraystretch}{2}
\centering
\resizebox{\linewidth}{!}{
\begin{tabular}{cccccc}
\hline
Model &$t$ &Dataset &$ACC$ &$ACC_{t}$  &$DP$ \\ \hline
\multirow{2}{*}{CuteGPT-13B} &\multirow{2}{*}{65\%} &C-Eval  &33.76\%  &\textbf{48.90}\%   &21.05\%  \\  
& &XieZhi &44.76\% &\textbf{73.62}\% &26.33\%   \\ 
\hline
\multirow{2}{*}{LLaMA2-13B} &\multirow{2}{*}{55\%} & GSM8K    & 30.67\%  &\textbf{46.3}\%   &24.30\%     \\ 
&   & CommonsenseQA         & 57.20\%    &\textbf{77.41}\%   & 38.42\%    \\ \hline
\end{tabular}
}
\caption{For CuteGPT-13B and LLaMA2-chat-13B, the accuracy performance under the different acceptable confidence thresholds on four datasets. $DP$ represents the proportion of data for which the confidence level exceeds the $t$.}
\label{table_2}
\end{table}
\textbf{Baselines.} We consider three different types of baseline approaches.

\emph{First token probability (First-Prob):} It uses the probability of the first token in the generation answer as the confidence score.

\emph{Subset clustering(SuC):} The dataset is initially partitioned into several subsets based on difficulty. The confidence score of each subset is computed through subset clustering. Ultimately, the LLM is trained using supervised fine-tuning. 

\emph{LePe-random:} This is a variant of our method. The key distinction lies in the utilization of random confidence during the learning stage, as opposed to the confidence computed by record D in our approach. \\
\textbf{Models.} We incorporate a range of widely used LLMs of different scales, including CuteGPT$\footnote{\url{https://github.com/Abbey4799/CuteGPT/}}$ and LLaMA2-Chat$\footnote{\url{https://huggingface.co/meta-llama/Llama-2-13b-chat/}}$. CuteGPT is built upon the original Llama model architecture, with expanded Chinese vocabulary and pretraining. It is available in two public versions: CuteGPT-7B and CuteGPT-13B. C-Eval and XieZhi are the Chinese datasets, therefore, we only employ CuteGPT on these two datasets. We use the 7B and 13B versions of the LLaMA2-Chat model on the GSM8K and CommonsenseQA. At the same time, we utilize the prompt used by Lin \citeyearpar{lin2022teaching} to guide ChatGPT to give its confidence level about the generation answer.\\
\textbf{Metrics.} To evaluate the accuracy of generated answers, we employ a string-matching approach to extract the model's final answer and compare them with the ground truth. We use the following evaluation metrics to assess the model's performance:\\
\textbf{\emph{Accuracy (ACC)}}. Represents the average accuracy of the LLM's responses.\\
\textbf{$ACC_{t}$}. Represents the accuracy of responses with confidence scores higher than the confidence threshold $t$.\\
\textbf{\emph{Expected Calibration Error (ECE)}}. We partition the inference results into $B$ disjoint bins based on the confidence scores, compute the average confidence score for each bin, and compare it with the average true accuracy of the answers within that bin. The ECE is calculated by:
\begin{equation}\label{eq:eq5}
ECE= \sum_{b=1}^{B}\frac{s_{b}}{S}\left | acc(b)-conf(b) \right | ,
\end{equation}
where $b$ is the $bth$ bin, $B$ is the total number of bins, $s_{b}$ is the number of questions in the $bth$ bin, $S$ is the total number of questions in the test set, $acc(b)$ is the true correctness of the answers in the $bth$ bin, and $conf(b)$ is the average of the LLM confidence in the $bth$ bin. The smaller the value, the better.\\
\textbf{\emph{Pearson Correlation Coefficient (r)}}. We also use the Pearson correlation coefficient to verify the correlation between the correctness of the LLM's responses and the confidence level after using our method :

\begin{tiny}
\begin{equation}\label{eq:eq6}
r=\frac{\sum_{b}^{B}(acc(b)-\overline{acc} )(conf(b)-\overline{conf} ) }{\sqrt{\sum_{b}^{B}(acc(b)-\overline{acc} )^{2}}  \sqrt{\sum_{b}^{B}(conf(b)-\overline{conf} )^{2}} },
\end{equation}
\end{tiny}
where $\overline{acc} $ is the average correctness of all questions, and $\overline{conf} $ is the average of the LLM confidence of all questions. \\
\textbf{Implementation Details.} Our optimizer is AdamW \cite{loshchilov2017decoupled} with $\beta _{1} $ and $\beta _{2} $ values of 0.9 and 0.95. During training, we set the initial learning rate to 1e-4, the final learning rate to 3e-4, the warmup phase to 300 steps, and we train for 700 steps. We conduct all our experiments using the NVIDIA A800.

\subsection{Experimental Analysis and Findings}
To validate the efficacy of our proposed method LePe, we mainly answer the following three questions.\\
\textbf{RQ1: Can our proposed LePe make LLM be calibrated?}\\
We conduct experiments to analyze the accuracy of LLM on the test data after fine-tuning our method. The overall result is shown in Figure \ref{fig:Result} and the quantitative results are presented in Table \ref{Result} and Figure \ref{fig:pineRes}.

\textbf{Overall analysis.} After using the LePe method, \textbf{we find that there is a strong correlation between the model's prediction confidence level and the actual accuracy}. For example, for LLaMA2-Chat-13B, the Pearson correlation coefficient between confidence and true correctness is 0.98 on the CommonsenseQA dataset.  For the LePe-random method, $r$ is only -0.20, there is almost no correlation between LLM confidence and true correctness to speak of. It also represents that our method can better capture LLM’s inherent true performance which helps LLM learn to express confidence. Additionally, LePe empowers the LLM with better calibration capability. We observe that the $ECE$ of our method is the lowest on almost all datasets. At the same time, the good calibration of our proposed method also is proved in the ACC metric. For example, CuteGPT-13B expresses 80-100\% confidence and exhibits a true correctness rate of 89.19\% on the XieZhi dataset. We also find for ChatGPT, the correlation coefficient is high, but the confidence level of its outputs is consistently high. For example, it barely outputs the confidence level between 0\%-40\% even if its answer is wrong. 

In our experimental datasets, we observe that for XieZhi and CommonsenseQA, whose inferential process is relatively simple, the CuteGPT obtains a high $r$ value and a low $ECE$ value. 
For instance, in the XieZhi dataset, when the LLM assigns a confidence range of 0-20\%, the actual correctness rate of its responses is about 25\%. It indicates that the calibration capability becomes significantly enhanced after using our proposed method. In contrast, with a more complexly structured reasoning dataset like GSM8K, the $ECE$ is relatively high. This shows that \textbf{the complexity of the posed questions impacts the calibration performance}.

\textbf{Parameter size.} For various parameter sizes, the calibration levels of LLM vary differently. However, it still follows the scaling laws \cite{kaplan2020scaling}. We find both CuteGPT and LLaMA2-Chat demonstrate superior calibration with the 13B parameters than the 7B parameters.  \\

\textbf{RQ2: Can the confidence level of the LLM output guide us to adopt the answer?}\\
For an LLM, we identify an acceptable confidence threshold through a comprehensive analysis of the confidence levels, the actual rate of correct responses, and the distribution of questions among different confidence intervals. The confidence threshold and the results are shown in Table \ref{table_2}. \textbf{The acceptable confidence threshold serves as a crucial guideline in practical applications.} If the generated confidence level of an LLM exceeds the set confidence threshold, we consider adopting its response. Otherwise, it means that the model is uncertain in answering the question, and we can consider using a larger language model to answer this question. In other words, this signal helps us discern the reliability of LLM's response. For example, for the CuteGPT-13B, we set the acceptable confidence threshold is 65\%. On the XieZhi dataset, 26.33\% of the questions displayed answer confidence levels surpassing the confidence threshold, and the corresponding correct response rate of 73.62\%, significantly surpassing the overall correct rate of 44.76\% on the whole dataset. Conversely, for ChatGPT, which is directly guided by the prompt to output confidence, the confidence is extremely high, and nearly no low-confidence responses. Hence, it is challenging for us to evaluate the answer based on the provided confidence levels.
\begin{table}[t]
\renewcommand{\arraystretch}{2}
\centering
\resizebox{\linewidth}{!}{
\begin{tabular}{ccccccc}
\hline
D-Train &D-Test &$t$ &$ACC$ &$ACC_{t}$  &$ECE$ \\ \hline
\multirow{2}{*}{XieZhi} &C-Eval &\multirow{2}{*}{65\%}   &33.76\%  &\textbf{38.74}\%    & 21.73  \\  
&M3KE & &27.05\% &\textbf{29.42}\%   &20.61   \\ 
\hline
\multirow{2}{*}{CommonsenseQA} &GSM8K &\multirow{2}{*}{55\%}   &30.61\%  &\textbf{33.62}\%  &22.19    \\  
&OpenBookQA & &41.32\% &\textbf{44.83}\%  &21.43\\ 
\hline
\end{tabular}
}
\caption{Testing on the out-of-domain datasets. The LLM is trained using LePe on D-Train and tested on D-Test.}
\label{table_3}
\end{table}

\textbf{RQ3: Can the LePe generalize to out-of-domain data?}\\
To assess the generalizability of our proposed method, we use CuteGPT-13B to train with LePe on XieZhi dataset and evaluate its confidence expression ability on C-Eval and M3KE \cite{liu2023m3ke}. Besides, we use LLaMA2-Chat-13B trained on CommonsenseQA and test its confidence expression ability on GSM8K and OpenBookQA \cite{mihaylov2018can}. The result is shown in Table \ref{table_3}.

\textbf{The calibration capability using LePe remains effective when tested with out-of-domain datasets}. For instance, after learning from the answer records in XieZhi, CuteGPT-13B still performs a good confidence expression on the C-Eval. When the model's output confidence score exceeds the acceptable confidence threshold, the accuracy rate reaches 38.74\%, surpassing the overall accuracy rate of 33.76\%. Additionally, GSM8K and CommonsenseQA are two completely different tasks, we find our method is still effective even when the test task varies significantly from the training task.
\section{Conclusion}
In this paper, we present a method of learning from past experience to enhance the LLMs’ capability for confidence expression, which enables the LLM to provide an answer and corresponding confidence. We first design a general pipeline to obtain the actual performance of the LLM on the problem. Further, we utilize the performance records to construct the dataset for instruction fine-tuning so that the LLM learns to express confidence in the generated answers. We conduct experiments on two open-source language LLMs to demonstrate the effectiveness of our method. The consistent experimental results across multiple tasks affirm that our method endows the LLM with confidence expression capability.
\section{Limitations}
We mainly evaluate LePe's performance in question-and-answer tasks and do not delve into its ability on open questions. Moreover, as mentioned earlier, using the confidence expressed by the model, we can identify the model's weaknesses and further improve them in a targeted manner, allowing LLM to continue to evolve. In this paper, we propose a general method to enhance the model's ability to express confidence, but we do not discuss the impact of ability on the continuous learning of large language models. 

\bibliography{LePe-ref}

\begin{thebibliography}{34}
\providecommand{\natexlab}[1]{#1}

\bibitem[{Achiam et~al.(2023)Achiam, Adler, Agarwal, Ahmad, Akkaya, Aleman, Almeida, Altenschmidt, Altman, Anadkat et~al.}]{achiam2023gpt}
Josh Achiam, Steven Adler, Sandhini Agarwal, Lama Ahmad, Ilge Akkaya, Florencia~Leoni Aleman, Diogo Almeida, Janko Altenschmidt, Sam Altman, Shyamal Anadkat, et~al. 2023.
\newblock Gpt-4 technical report.
\newblock \emph{arXiv preprint arXiv:2303.08774}.

\bibitem[{Branwen(2020)}]{Branwen}
Gwern Branwen. 2020.
\newblock \href {https://gwern.net/gpt-3-nonfiction#calibration} {Gpt-3 nonfiction- calibration}.
\newblock Technical report, The institution that published.
\newblock Last accessed on 2022-04-24.

\bibitem[{Cobbe et~al.(2021)Cobbe, Kosaraju, Bavarian, Chen, Jun, Kaiser, Plappert, Tworek, Hilton, Nakano et~al.}]{cobbe2021gsm8k}
Karl Cobbe, Vineet Kosaraju, Mohammad Bavarian, Mark Chen, Heewoo Jun, Lukasz Kaiser, Matthias Plappert, Jerry Tworek, Jacob Hilton, Reiichiro Nakano, et~al. 2021.
\newblock Training verifiers to solve math word problems.
\newblock \emph{arXiv preprint arXiv:2110.14168}.

\bibitem[{Gal and Ghahramani(2016)}]{gal2016dropout}
Yarin Gal and Zoubin Ghahramani. 2016.
\newblock Dropout as a bayesian approximation: Representing model uncertainty in deep learning.
\newblock In \emph{international conference on machine learning}, pages 1050--1059. PMLR.

\bibitem[{Giallanza and Campbell(2024)}]{giallanza2024contextsensitive}
Tyler Giallanza and Declan~Iain Campbell. 2024.
\newblock Context-sensitive semantic reasoning in large language models.
\newblock In \emph{ICLR 2024 Workshop on Representational Alignment}.

\bibitem[{Gu et~al.(2024)Gu, Zhu, Ye, Zhang, Wang, Zhu, Jiang, Xiong, Li, Wu et~al.}]{gu2024xiezhi}
Zhouhong Gu, Xiaoxuan Zhu, Haoning Ye, Lin Zhang, Jianchen Wang, Yixin Zhu, Sihang Jiang, Zhuozhi Xiong, Zihan Li, Weijie Wu, et~al. 2024.
\newblock Xiezhi: An ever-updating benchmark for holistic domain knowledge evaluation.
\newblock In \emph{Proceedings of the AAAI Conference on Artificial Intelligence}, volume~38, pages 18099--18107.

\bibitem[{Guo et~al.(2023)Guo, Zhang, Wang, Jiang, Nie, Ding, Yue, and Wu}]{guo2023close}
Biyang Guo, Xin Zhang, Ziyuan Wang, Minqi Jiang, Jinran Nie, Yuxuan Ding, Jianwei Yue, and Yupeng Wu. 2023.
\newblock How close is chatgpt to human experts? comparison corpus, evaluation, and detection.
\newblock \emph{arXiv preprint arXiv:2301.07597}.

\bibitem[{Han et~al.(2024)Han, Liang, Shi, He, and Xiao}]{Han2024SmallLM}
Haixia Han, Jiaqing Liang, Jie Shi, Qi~He, and Yanghua Xiao. 2024.
\newblock \href {https://api.semanticscholar.org/CorpusID:266999677} {Small language model can self-correct}.
\newblock In \emph{AAAI Conference on Artificial Intelligence}.

\bibitem[{Huang et~al.(2023)Huang, Yu, Ma, Zhong, Feng, Wang, Chen, Peng, Feng, Qin, and Liu}]{Huang2023ASO}
Lei Huang, Weijiang Yu, Weitao Ma, Weihong Zhong, Zhangyin Feng, Haotian Wang, Qianglong Chen, Weihua Peng, Xiaocheng Feng, Bing Qin, and Ting Liu. 2023.
\newblock \href {https://api.semanticscholar.org/CorpusID:265067168} {A survey on hallucination in large language models: Principles, taxonomy, challenges, and open questions}.
\newblock \emph{ArXiv}, abs/2311.05232.

\bibitem[{Huang et~al.(2024)Huang, Bai, Zhu, Zhang, Zhang, Su, Liu, Lv, Zhang, Fu et~al.}]{huang2024ceval}
Yuzhen Huang, Yuzhuo Bai, Zhihao Zhu, Junlei Zhang, Jinghan Zhang, Tangjun Su, Junteng Liu, Chuancheng Lv, Yikai Zhang, Yao Fu, et~al. 2024.
\newblock C-eval: A multi-level multi-discipline chinese evaluation suite for foundation models.
\newblock \emph{Advances in Neural Information Processing Systems}, 36.

\bibitem[{Kadavath et~al.(2022)Kadavath, Conerly, Askell, Henighan, Drain, Perez, Schiefer, Dodds, DasSarma, Tran-Johnson, Johnston, El-Showk, Jones, Elhage, Hume, Chen, Bai, Bowman, Fort, Ganguli, Hernandez, Jacobson, Kernion, Kravec, Lovitt, Ndousse, Olsson, Ringer, Amodei, Brown, Clark, Joseph, Mann, McCandlish, Olah, and Kaplan}]{Kadavath2022LanguageM}
Saurav Kadavath, Tom Conerly, Amanda Askell, Tom Henighan, Dawn Drain, Ethan Perez, Nicholas Schiefer, Zachary Dodds, Nova DasSarma, Eli Tran-Johnson, Scott Johnston, Sheer El-Showk, Andy Jones, Nelson Elhage, Tristan Hume, Anna Chen, Yuntao Bai, Sam Bowman, Stanislav Fort, Deep Ganguli, Danny Hernandez, Josh Jacobson, John Kernion, Shauna Kravec, Liane Lovitt, Kamal Ndousse, Catherine Olsson, Sam Ringer, Dario Amodei, Tom~B. Brown, Jack Clark, Nicholas Joseph, Benjamin Mann, Sam McCandlish, Christopher Olah, and Jared Kaplan. 2022.
\newblock \href {https://api.semanticscholar.org/CorpusID:250451161} {Language models (mostly) know what they know}.
\newblock \emph{ArXiv}, abs/2207.05221.

\bibitem[{Kaplan et~al.(2020)Kaplan, McCandlish, Henighan, Brown, Chess, Child, Gray, Radford, Wu, and Amodei}]{kaplan2020scaling}
Jared Kaplan, Sam McCandlish, Tom Henighan, Tom~B Brown, Benjamin Chess, Rewon Child, Scott Gray, Alec Radford, Jeffrey Wu, and Dario Amodei. 2020.
\newblock Scaling laws for neural language models.
\newblock \emph{arXiv preprint arXiv:2001.08361}.

\bibitem[{Li et~al.(2024{\natexlab{a}})Li, Patel, Vi{\'e}gas, Pfister, and Wattenberg}]{li2024inference}
Kenneth Li, Oam Patel, Fernanda Vi{\'e}gas, Hanspeter Pfister, and Martin Wattenberg. 2024{\natexlab{a}}.
\newblock Inference-time intervention: Eliciting truthful answers from a language model.
\newblock \emph{Advances in Neural Information Processing Systems}, 36.

\bibitem[{Li et~al.(2024{\natexlab{b}})Li, Wang, Feng, Zhu, Wang, and Chua}]{li2024think}
Moxin Li, Wenjie Wang, Fuli Feng, Fengbin Zhu, Qifan Wang, and Tat-Seng Chua. 2024{\natexlab{b}}.
\newblock Think twice before assure: Confidence estimation for large language models through reflection on multiple answers.
\newblock \emph{arXiv preprint arXiv:2403.09972}.

\bibitem[{Lin et~al.(2022)Lin, Hilton, and Evans}]{lin2022teaching}
Stephanie Lin, Jacob Hilton, and Owain Evans. 2022.
\newblock Teaching models to express their uncertainty in words.
\newblock \emph{arXiv preprint arXiv:2205.14334}.

\bibitem[{Liu et~al.(2023)Liu, Jin, Ren, Yu, Dong, Peng, Zhang, Peng, Zhang, Lyu et~al.}]{liu2023m3ke}
Chuang Liu, Renren Jin, Yuqi Ren, Linhao Yu, Tianyu Dong, Xiaohan Peng, Shuting Zhang, Jianxiang Peng, Peiyi Zhang, Qingqing Lyu, et~al. 2023.
\newblock M3ke: A massive multi-level multi-subject knowledge evaluation benchmark for chinese large language models.
\newblock \emph{arXiv preprint arXiv:2305.10263}.

\bibitem[{Liu et~al.(2024)Liu, Zhuang, Bi, Huang, Huang, Li, Yu, Liu, Hu, Hong et~al.}]{liu2024survey}
Qi~Liu, Yan Zhuang, Haoyang Bi, Zhenya Huang, Weizhe Huang, Jiatong Li, Junhao Yu, Zirui Liu, Zirui Hu, Yuting Hong, et~al. 2024.
\newblock Survey of computerized adaptive testing: A machine learning perspective.
\newblock \emph{arXiv preprint arXiv:2404.00712}.

\bibitem[{Loshchilov and Hutter(2017)}]{loshchilov2017decoupled}
Ilya Loshchilov and Frank Hutter. 2017.
\newblock Decoupled weight decay regularization.
\newblock \emph{arXiv preprint arXiv:1711.05101}.

\bibitem[{Mihaylov et~al.(2018)Mihaylov, Clark, Khot, and Sabharwal}]{mihaylov2018can}
Todor Mihaylov, Peter Clark, Tushar Khot, and Ashish Sabharwal. 2018.
\newblock Can a suit of armor conduct electricity? a new dataset for open book question answering.
\newblock \emph{arXiv preprint arXiv:1809.02789}.

\bibitem[{Ouyang et~al.(2022)Ouyang, Wu, Jiang, Almeida, Wainwright, Mishkin, Zhang, Agarwal, Slama, Ray et~al.}]{ouyang2022training}
Long Ouyang, Jeffrey Wu, Xu~Jiang, Diogo Almeida, Carroll Wainwright, Pamela Mishkin, Chong Zhang, Sandhini Agarwal, Katarina Slama, Alex Ray, et~al. 2022.
\newblock Training language models to follow instructions with human feedback.
\newblock \emph{Advances in neural information processing systems}, 35:27730--27744.

\bibitem[{Rafailov et~al.(2024)Rafailov, Sharma, Mitchell, Manning, Ermon, and Finn}]{rafailov2024direct}
Rafael Rafailov, Archit Sharma, Eric Mitchell, Christopher~D Manning, Stefano Ermon, and Chelsea Finn. 2024.
\newblock Direct preference optimization: Your language model is secretly a reward model.
\newblock \emph{Advances in Neural Information Processing Systems}, 36.

\bibitem[{Ren et~al.(2023)Ren, Wang, Qu, Zhao, Liu, Tian, Wu, Wen, and Wang}]{ren2023investigating}
Ruiyang Ren, Yuhao Wang, Yingqi Qu, Wayne~Xin Zhao, Jing Liu, Hao Tian, Hua Wu, Ji-Rong Wen, and Haifeng Wang. 2023.
\newblock Investigating the factual knowledge boundary of large language models with retrieval augmentation.
\newblock \emph{arXiv preprint arXiv:2307.11019}.

\bibitem[{Shuster et~al.(2021)Shuster, Poff, Chen, Kiela, and Weston}]{shuster2021retrieval}
Kurt Shuster, Spencer Poff, Moya Chen, Douwe Kiela, and Jason Weston. 2021.
\newblock Retrieval augmentation reduces hallucination in conversation.
\newblock \emph{arXiv preprint arXiv:2104.07567}.

\bibitem[{Srivastava et~al.(2022)Srivastava, Rastogi, Rao, Shoeb, Abid, Fisch, Brown, Santoro, Gupta, Garriga-Alonso et~al.}]{srivastava2022beyond}
Aarohi Srivastava, Abhinav Rastogi, Abhishek Rao, Abu Awal~Md Shoeb, Abubakar Abid, Adam Fisch, Adam~R Brown, Adam Santoro, Aditya Gupta, Adri{\`a} Garriga-Alonso, et~al. 2022.
\newblock Beyond the imitation game: Quantifying and extrapolating the capabilities of language models.
\newblock \emph{arXiv preprint arXiv:2206.04615}.

\bibitem[{Stiennon et~al.(2020)Stiennon, Ouyang, Wu, Ziegler, Lowe, Voss, Radford, Amodei, and Christiano}]{stiennon2020learning}
Nisan Stiennon, Long Ouyang, Jeffrey Wu, Daniel Ziegler, Ryan Lowe, Chelsea Voss, Alec Radford, Dario Amodei, and Paul~F Christiano. 2020.
\newblock Learning to summarize with human feedback.
\newblock \emph{Advances in Neural Information Processing Systems}, 33:3008--3021.

\bibitem[{Talmor et~al.(2018)Talmor, Herzig, Lourie, and Berant}]{talmor2018commonsenseqa}
Alon Talmor, Jonathan Herzig, Nicholas Lourie, and Jonathan Berant. 2018.
\newblock Commonsenseqa: A question answering challenge targeting commonsense knowledge.
\newblock \emph{arXiv preprint arXiv:1811.00937}.

\bibitem[{Tian et~al.(2023{\natexlab{a}})Tian, Mitchell, Yao, Manning, and Finn}]{tian2023fine}
Katherine Tian, Eric Mitchell, Huaxiu Yao, Christopher~D Manning, and Chelsea Finn. 2023{\natexlab{a}}.
\newblock Fine-tuning language models for factuality.
\newblock \emph{arXiv preprint arXiv:2311.08401}.

\bibitem[{Tian et~al.(2023{\natexlab{b}})Tian, Mitchell, Zhou, Sharma, Rafailov, Yao, Finn, and Manning}]{tian2023just}
Katherine Tian, Eric Mitchell, Allan Zhou, Archit Sharma, Rafael Rafailov, Huaxiu Yao, Chelsea Finn, and Christopher~D Manning. 2023{\natexlab{b}}.
\newblock Just ask for calibration: Strategies for eliciting calibrated confidence scores from language models fine-tuned with human feedback.
\newblock \emph{arXiv preprint arXiv:2305.14975}.

\bibitem[{Wang et~al.(2024)Wang, Ma, Hu, Weber-Genzel, R{\"o}ttger, Kreuter, Hovy, and Plank}]{Wang2024MyAI}
Xinpeng Wang, Bolei Ma, Chengzhi Hu, Leon Weber-Genzel, Paul R{\"o}ttger, Frauke Kreuter, Dirk Hovy, and Barbara Plank. 2024.
\newblock \href {https://api.semanticscholar.org/CorpusID:267782369} {"my answer is c": First-token probabilities do not match text answers in instruction-tuned language models}.
\newblock \emph{ArXiv}, abs/2402.14499.

\bibitem[{Wang et~al.(2022)Wang, Wei, Schuurmans, Le, Chi, Narang, Chowdhery, and Zhou}]{wang2022self}
Xuezhi Wang, Jason Wei, Dale Schuurmans, Quoc Le, Ed~Chi, Sharan Narang, Aakanksha Chowdhery, and Denny Zhou. 2022.
\newblock Self-consistency improves chain of thought reasoning in language models.
\newblock \emph{arXiv preprint arXiv:2203.11171}.

\bibitem[{Wei et~al.(2022)Wei, Wang, Schuurmans, Bosma, hsin Chi, Xia, Le, and Zhou}]{Wei2022ChainOT}
Jason Wei, Xuezhi Wang, Dale Schuurmans, Maarten Bosma, Ed~Huai hsin Chi, F.~Xia, Quoc Le, and Denny Zhou. 2022.
\newblock \href {https://api.semanticscholar.org/CorpusID:246411621} {Chain of thought prompting elicits reasoning in large language models}.
\newblock \emph{ArXiv}, abs/2201.11903.

\bibitem[{Xiong et~al.(2023)Xiong, Hu, Lu, Li, Fu, He, and Hooi}]{xiong2023can}
Miao Xiong, Zhiyuan Hu, Xinyang Lu, Yifei Li, Jie Fu, Junxian He, and Bryan Hooi. 2023.
\newblock Can llms express their uncertainty? an empirical evaluation of confidence elicitation in llms.
\newblock \emph{arXiv preprint arXiv:2306.13063}.

\bibitem[{Yin et~al.(2023)Yin, Sun, Guo, Wu, Qiu, and Huang}]{yin2023large}
Zhangyue Yin, Qiushi Sun, Qipeng Guo, Jiawen Wu, Xipeng Qiu, and Xuanjing Huang. 2023.
\newblock Do large language models know what they don't know?
\newblock \emph{arXiv preprint arXiv:2305.18153}.

\bibitem[{Zhou et~al.(2023)Zhou, Jurafsky, and Hashimoto}]{zhou2023navigating}
Kaitlyn Zhou, Dan Jurafsky, and Tatsunori Hashimoto. 2023.
\newblock Navigating the grey area: How expressions of uncertainty and overconfidence affect language models.
\newblock \emph{arXiv preprint arXiv:2302.13439}.

\end{thebibliography}
\end{document}